\def\year{2019}\relax
\documentclass[letterpaper]{article} %
\usepackage{aaai19}  %
\usepackage{times}  %
\usepackage{helvet}  %
\usepackage{courier}  %
\usepackage{url}  %
\usepackage{graphicx}  %
\usepackage[utf8]{inputenc}

\usepackage{wrapfig}
\usepackage{textcomp}
\usepackage[table,dvipsnames]{xcolor}
\usepackage{epsfig}
\usepackage{pgfplotstable}
\usepackage{pgfplots}
\usepgfplotslibrary{groupplots}
\usepackage{amsmath}
\usepackage{amssymb}
\usepackage{bbm}
\usepackage{float}
\usepackage{hyperref}
\usepackage{microtype}
\usepackage{tikz}
\usetikzlibrary{decorations.text,calc,shapes,arrows,arrows.meta, positioning,shapes.misc,decorations.markings,decorations.markings,decorations.pathreplacing,matrix}
\usepackage{booktabs}
\usepackage{multirow}
\usepackage{adjustbox}
\usepackage{verbatim}
\usepackage[T1]{fontenc}
\usepackage[title]{appendix}
\usepackage{adjustbox} %
\usepackage{graphicx}
\usepackage{caption}
\usepackage{subcaption}
\usepackage{algpseudocode}
\usepackage{algorithm}
\usepackage{siunitx}
\usepackage{nicefrac}
\usepackage[colorinlistoftodos,textsize=footnotesize]{todonotes}
\usepackage{bm}
\usepackage{xspace}

\usepackage[]{natbib}

\usepackage[autostyle, english=american]{csquotes}
\MakeOuterQuote{"}

\makeatletter
\pgfplotsset{
    every axis x label/.append style={
        alias=current axis xlabel
    },
    legend pos/outer south/.style={
        /pgfplots/legend style={
            at={%
                (%
                \@ifundefined{pgf@sh@ns@current axis xlabel}%
                {xticklabel cs:0.5}%
                {current axis xlabel.south}%
                )%
            },
            anchor=north
        }
    }
}
\makeatother

\newcolumntype{t}{>{\ttfamily}l}
\newcolumntype{T}{>{\ttfamily}c}

\newcolumntype{$}{>{\global\let\currentrowstyle\relax}}
\newcolumntype{^}{>{\currentrowstyle}}

\DeclareMathOperator*{\sign}{sign}

\usepackage{setspace}

\hypersetup{draft}

\def\malconvP{MalConv\textsuperscript{+}\xspace}
\def\ngramP{N-Gram\textsuperscript{+}\xspace}

\title{\Large \bf Non-Negative Networks Against Adversarial Attacks}

\author{
  William Fleshman,\textsuperscript{1} Edward Raff,\textsuperscript{1,2} Jared Sylvester,\textsuperscript{1,2} Steven Forsyth,\textsuperscript{3} Mark McLean\textsuperscript{1} \\
  \textsuperscript{1}Laboratory for Physical Sciences, \textsuperscript{2}Booz Allen Hamilton, \textsuperscript{3}Nvidia  \\
  \texttt{\{william.fleshman, edraff, jared, mrmclea\}@lps.umd.edu}, \texttt{sforsyth@nvidia.com}
}

\begin{document}

\maketitle

\begin{abstract}
Adversarial attacks against neural networks are a problem of considerable importance, for which effective defenses are not yet readily available. We make progress toward this problem by showing that non-negative weight constraints can be used to improve resistance in specific scenarios. In particular, we show that they can provide an effective defense for binary classification problems with asymmetric cost, such as malware or spam detection. 
We also show the potential for non-negativity to be helpful to non-binary problems by applying it to image classification. 
\end{abstract}

\section{Introduction}
Recently, there has been an increased research effort in exploring adversarial examples which fool machine learning  classifiers~\citep{Goodfellow2015, Kurakin2017AdversarialWorld,  Szegedy2014,Yuan2017}. The majority of the existing research focuses on the image domain, where an example is generated by making small perturbations to input pixels in order to make a large change in the distribution of predicted class probabilities. We are particularly interested in adversarial attacks for \textit{malware detection}, which is the task of determining if a file is benign or malicious. This involves a real-life adversary (the malware author) who is attempting to subvert detection tools, such as anti-virus programs.
With machine learning approaches to malware detection becoming more prevalent \citep{MalConv, export:249072, Saxe2015, Sahs2012}, this is an area that urgently requires solutions to the adversarial problem.
Because an adversary is actively attempting to subvert outputs, small decreases in accuracy when not under attack are an acceptable cost for remediating targeted attacks. In this scenario, the effective accuracy of the system would be the accuracy under attack, which will be at or near zero without proper defenses.

For example, \citet{MalConv} trained a convolutional neural network called MalConv to distinguish between benign and malicious Windows executable files. 
When working with images, any pixel can be arbitrarily altered, but this freedom does not carry over to the malware case. The executable format follows stricter rules which constrain the options available to the attacker \citep{Kreuk2018,Russu:2016:SKM:2996758.2996771,DBLP:journals/corr/GrossePM0M16,Suciu2018}. 
Perturbing an arbitrary byte of an executable file will most likely change the functionality of the file or prevent it from executing entirely. This property is useful for defending against an adversarial attack, as a malware author needs to evade detection with a \textit{working} malicious file.

\citet{Kreuk2018} were able to bypass these limitations by applying gradient-based attacks to create perturbations which were restricted to bytes located in unused sections of malicious executable files. The adversarial examples remained just as malicious, but the classifier was fooled by the introduction of overwhelmingly benign yet unused sections of the file.
This is possible because the adversary controls the input, 
and the EXE format allows unused sections.  
Because of the complications and obfuscations that are available to malware authors, it is not necessarily possible to tell that a section is unused,
even if its contents appear random. This is an \textit{additive only} adversary --- i.e., the attacker can only add features --- which has been widely used and will be the focus of our study. 

An analogy to the image domain would be an attacker that could create new pixels which represent the desired class and put them outside of the cropping box of the image, such that they would be in the digital file, but never be seen by a human observer. This contrasts with a standard adversarial attack on images, since the attacker is typically limited to changing the values of existing pixels in the image rather than introducing new pixels entirely.

Given these unique characteristics and costs, we note that the malware case is one where we care \textit{only} about targeted adversarial attacks. The adversary always wants to fool detectors into calling malicious files benign. As such, we introduce an approach to tackle targeted adversarial attacks by exploiting non-negative learning constraints. We will highlight related work in \autoref{sec:related}. In \autoref{sec:method} we will detail our motivation for non-negative learning for malware, as well as how we generalize its use to multi-class problems like image classifiers. The attack scenario and experiments on malware, spam, and image domains will be detailed in \autoref{sec:experiments}. In \autoref{sec:results} we will demonstrate how our approach reduces evasions to almost 0\% for malware and exactly 0\% spam detection. On images we show improvements to robustness against confident adversarial attacks against images, showing that there is potential for non-negativity to aid in non-binary problems. We will end with our conclusions in \autoref{sec:conclusion}.

\section{Related Work} \label{sec:related}

The issues of targeted adversarial binary classification problems, as well as the additive adversary, was first brought up by \citet{Dalvi:2004:AC:1014052.1014066}, who noted its importance in a number of domains like fraud detection, counter terrorism, surveillance, and others. There have been several attempts at creating machine learning classifiers which can defend against such adversarial examples. \citet{Yuan2017} provide a thorough survey of both attacks and defenses specifically for deep learning systems. Some of these attacks will be used to compare the robustness of our technique to prior methods.

In our case we are learning against a real life adversary in a binary classification task, similar to the initial work in this space on evading spam filters \citep{Lowd2005a,Dalvi:2004:AC:1014052.1014066,Lowd:2005:AL:1081870.1081950}. Our malware case gives the defender a slight comparative advantage in constraining the attack to produce a working binary, where spam authors can insert more arbitrary content. 

Prior works have looked at similar weight constraint approaches to adversarial robustness. \citet{citeulike:7099488} uses a technique to keep the distribution of learned weights associated with features as even as possible during training. By preventing any one feature from becoming overwhelmingly predictive, they force the adversary to manipulate many features in order to cause a misclassification. 
Similarly, \citet{DBLP:journals/corr/GrossePM0M16} tested a suite of feature reduction methods specifically in the malware domain~\citep{DBLP:journals/corr/GrossePM0M16}. First, they used the mutual information between features and the target class in order to limit the representation of each file to those features. Like \citeauthor{citeulike:7099488}, they created an alternative feature selection method to limit training to features which carried near equal importance. They found both of these techniques to be ineffective. 

Our approach is also a feature reduction technique. The difference is that we train on all features, but only retain the capacity to distinguish a reduced number of features at test time --- namely, only those indicative of the positive class. Training on all features allows the model to automatically determine which are important for the target class and utilizes the other features to accurately set a threshold, represented by the bias term, for determining when a requisite quantity of features are present for assigning samples to the target class.

\citet{Chorowski2015} used non-negative weight constraints in order to train more interpretable neural networks. They found that the constraints caused the neurons to isolate features in meaningful ways. We build on this technique in order to isolate features while also preventing our models from using the features predictive of the negative class.

\citet{Goodfellow2015} used RBF networks to show that low capacity models can be robust to adversarial perturbations but found they lack the ability to generalize. With our methods we find we are able to achieve generalization while also producing low confidence predictions during targeted attacks.

\section{Isolating Classes with Non-Negative Weight Constraints} \label{sec:method}

We will start by building an intuition on how logistic regression with non-negative weight constraints assigns predictive power to only features indicative of the positive ($+$) class while ignoring those associated with the negative ($-$) class.

Let $\bm{C}(\cdot)$ be a trained logistic regression binary classifier of the form $\bm{C}(\bm{x}) = \sign \left( \bm{w}^{\mathsf{T}}\bm{x} + b \right)$,
where $\bm{w}$ is the vector of non-negative learned coefficients of $\bm{C}(\cdot)$, $\bm{x}$ is a vector of boolean features for a given sample, and $b$ is a scalar bias. The decision boundary of $\bm{C}(\cdot)$ exists where $\bm{w}^{\mathsf{T}}\bm{x}+b = 0$, and because $\bm{w}^{\mathsf{T}}\bm{x} \geq 0$ $ \forall $ $\bm{x}$, the bias $b$ must be strictly negative in order for $\bm{C}(\cdot)$ to have the capacity to assign samples to both classes. The decision function can then be rewritten as:
\begin{equation}\label{eq:logreg_constrain}
	\bm{C}(\bm{x}) =
   \begin{cases} 
      (+) & \bm{w}^{\mathsf{T}}\bm{x} \geq |b| \\
      (-) & \bm{w}^{\mathsf{T}}\bm{x} < |b| 
   \end{cases}
\end{equation}
Because $\bm{w}$ is non-negative, the presence of any feature $x_i \in \bm{x}$ can only increase the result of the dot product, thus pushing the classification toward $(+)$. Weights associated to features that are predictive of class $(-)$ will therefore be pushed toward $0$ during training. When no features are present $(\bm{x} = \vec{0})$ the  model defaults to a classification of $(-)$ due to the negative bias $b$. Unless a sufficient number of features predictive of class $(+)$ are present in the sample, the decision will remain unchanged. A classifier trained in this way will use features indicative of the $(-)$ class to set the bias term, but will not allow those features to participate in classification at test time. The same logic follows for logistic regression with non-boolean features if the features are also non-negative or scaled to be non-negative before training. 

Given a problem with asymmetric misclassification goals, we can leverage this behavior to build a defense against adversarial attacks. For malware detection, the malware author wishes to avoid detection as malware $(+)$, and instead induce a false detection as benign $(-)$. However, there is no desire for the author of a benign program to make their applications be detected as malicious. Thus, if we model malware as the positive class with non-negative weights, \textit{nothing can be added} to the file to make it seem more benign to the classifier $\bm{C}(\cdot)$. Because executable programs must maintain functionality, the malware author can not trivially remove content to reduce the malicious score either.  This leaves the attacker with no recourse but to re-write their application, or perform more non-trivial acts such as packing to obscure information. Such obfuscations can then be remediated through existing approaches like dynamic analysis \citep{Ugarte-Pedrero:2016:RRP:2976956.2976970,Chistyakov2017}. 

Notably, this method also applies to neural networks with a sigmoid output neuron as long as the input to the final layer and the final layer's weights are constrained to be non-negative. The output layer of such a network is identical to our logistic regression example. The cumulative operation of the intermediate layers $\bm{\phi}(\cdot)$ can be interpreted as a re-representation of the features before applying the logistic regression such as $\bm{C}(\bm{x}) = \sign \left( \bm{w}^{\mathsf{T}}\bm{\phi}(\bm{x}) + b \right)$.
We will denote when a model is trained in a non-negative fashion by appending "\hspace{1pt}\textsuperscript{+}\hspace{1pt}" to its name. 

The ReLU function is a good choice for intermediate layers as it maintains the required non-negative representation and is already found in most modern neural networks. 

\begin{figure}[!h]
\vspace{0.25\baselineskip}
\centering
\includegraphics[width=1.0\columnwidth]{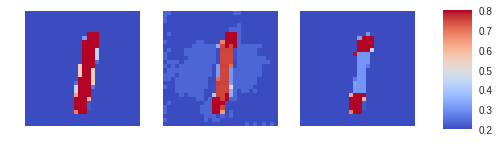}
\caption{Left: Original Image; Middle: Gradient attack on LeNet; Right: Gradient attack on non-negative LeNet\textsuperscript{+}. The attack on the standard model was able to add pixel intensity in a round, zero-shaped area to fool the classifier into thinking this was a zero. The attack on the constrained model was forced to remove pixel intensity from the one rather than adding in new values elsewhere.}
\label{fig:binary_mnist_gradient}
\end{figure}

For building intuition, in \autoref{fig:binary_mnist_gradient} we provide an example of how this works for neural networks using MNIST. To fool the network into predicting the positive class (one) as the negative class (zero), the adversary must now make larger removals of content --- to the point that the non-negative attack is no longer a realistic input.  

It should be noted that constraining a model in this way does reduce the amount of information available for discriminating samples at inference time, and a drop in classification accuracy is likely to occur for most problems. The trade off between adversarial robustness and performance should be analyzed for the specific domain and use case. 

A practical benefit to our approach is that it is simple to implement. In the  general case, on can simply use gradient decent with a projection step that clips negative values to zero after each step update. We implemented our approach in Keras \citep{chollet2015keras} by simply adding the "NonNeg" constraint to each layer in the model.\footnote{\url{https://keras.io/constraints/}} 

\subsection{Non-Negativity and Multi-Class Classification} \label{sec:image}

The primary focus of our work is on binary tasks like malware and spam detection, it is also worth asking if it can be applied to multi-class problems. 
In this work we show that non-negativity can still have some benefit in this scenario, but we find it necessary to re-phrase how such tasks are handled. Normally, one would use the softmax 
($\text{softmax}(\bm{v})_i = {\exp(v_i)}/{\sum_{j=1}^{n} \exp(v_j)}$)
on the un-normalized probabilities $\bm{v}$ given by the final layer. The probability of a class $i$ is then taken as $\text{softmax}(\bm{v})_i$. However we find that the softmax activation makes it easier to attack networks.

Take the non attacked activation pattern $\bm{v}$, where $v_i > v_j$ $ \forall $ $j \neq i$. Now consider the new activation pattern $\bm{\hat{v}}$, which is produced by an adversarially perturbed input with the goal of inducing a prediction as class $q$ instead of $i$. Then it is necessary to force $\hat{v}_q > \hat{v}_i$. Yet even if $\hat{v}_i \approx v_i$, the probability of class $i$ can be made arbitrarily low by continuing to maximize the response of $\hat{\bm{v}}_q$. This means we are able to diminish the apparent probability of class $i$ without having impacted the model's response to class $i$. Phrased analogously as an image classification problem, adversaries don't need to remove the amount of "cat" in a photo to induce a decision of "potato," but only increase the amount of "potato." 

In addition \citet{Chorowski2015} proved that a non-negative network trained with softmax activation can be transformed into an equivalent unconstrained network. This means there is little reason to expect our non-negative approach to provide benefit if we stay with the softmax activation, as it has an equivalent unconstrained form and should be equally susceptible to all adversarial attacks.  As such we must move away from softmax to get the benefits of our non-negative approach in a multi-class scenario. 

Instead we can look at the classification problem in a one-vs-all fashion by replacing the softmax activation over $K$ classes with $K$ independent classifications trained with the binary cross-entropy loss and using the sigmoid activation $\sigma(z) = 1/(1+\exp(-z))$. Final probabilities after training are obtained by normalizing the sigmoid responses to sum to one. We find that this strategy combined with non-negative learning provides some robustness against an adversary producing targeted high-confidence attacks (e.g., the network is 99\% sure the cat is a potato). The one-vs-all component make it such that increasing the confidence of a new class eventually requires reducing the confidence of the original class. The non-negativity increases the difficulty of this removal step, resulting in destructive changes to the image. 

We make two important notes on how we apply non-negative training for image classification. First, we pre-train the network using the standard softmax activation, and then re-train the weights with our one-vs-all style and non-negative constraints on the final fully connected layers. Doing so we find only a small difference in accuracy between results, where training non-negative networks from scratch often has reduced accuracy. Second, we continue to use batch normalization without constraints. This is because batch normalization can be rolled into the bias term and as a re-scaling of the weights, and so does not break the non-negative constraint in any way. We find its positive impact on convergence greater when training with the non-negative constraints.

\section{Experimental Methodology} \label{sec:experiments}

Having defined the mechanism by which we will defend against targeted adversarial attacks, we will investigate its application to two malware detection models, one spam detection task, and four image classification tasks. 
We will spend more time introducing the malware attacks, as readers may not have as much experience with this domain.

For malware, we will look at MalConv \citep{MalConv}, a recently proposed neural network that learns from raw bytes. We will also consider an N-Gram based model~\citep{raff_ngram_2016}. Both of these techniques are applied to the raw bytes of a file. We use the same 2,000,000 training and 80,000 testing datums as used in \citet{raff_shwel}.

Following recommendations by \citet{Biggio2014} we will specify the threat model under which we perform our evaluations. In all cases, our threat model will assume a white-box adversary that has full knowledge of the models, their weights, and the training data. For our binary classification problems, we assume in the threat model that our adversary can only add new features to the model (i.e., in the feature vector space they can change a zero valued feature to non-zero, but can not alter an already non-zero value). 

We recognize that this threat model does not encompass all possible adversaries, but note that it is one of the most commonly used adversarial models spanning many domains. The "Good Word" attack on spam messages is itself an example of this threat model's action space, and one of the initial works in adversarial learning noted its wide applicability \citep{Lowd:2005:AL:1081870.1081950}. In a recent survey, \citet{Maiorca2018} found that 9 out of 10 works in evading malicious PDF detectors used the additive only threat model, and these additive adversaries succeeded in both white-box and black-box attacks. \citet{Demontis2017} considered both the additive only adversary, as well as one which could add or remove features, as applied to android malware detection. On their Android data, they demonstrate a learning approach which provides bounds on adversary success under both adversary action models, making it robust but still vulnerable. Under the white-box additive attack scenario, their Secure-SVM detection rate drops from 95\% on normal test data down to 60\% when attacked. Finally, for the case of Windows PE data, three different works have attacked the MalConv model using the additive adversary~\citep{Kreuk2018,Kolosnjaji2018,Suciu2018}. 

The additive threat model makes  sense to study, as it is easier to implement for the adversary and currently successful in practice. For this reason, it makes little sense for the adversary to consider a more powerful threat model (e.g., adding and removing features) which would increase their costs and effort, when the simpler and cheaper alternative works. We will show in \autoref{sec:results} that while not perfect, our non-negative defense is the first to demonstrably thwart the additive adversary while still obtaining reasonable accuracies. This forces a potential adversary to "step up" to a more powerful model, which increases their effort and cost. We contend this is of intrinsic value \textit{eo ipso}. Below we will review additional details regarding the threat model for each data-type we consider (Windows PE, emails, and images). This is followed by specifics on how the attacks are carried out for each classification algorithm as the details are different in all cases due to model and problem diversity. 

\paragraph{Windows PE Threat Model Specifics}

For PE malware we use the appending of an unused section as the attack vector for technical simplicity. The adversary will be allowed to append any desired number of bytes into an added and unused section of the binary file, until no change in the evasion rate occurs. Our approach should still work if the adversary performed insertions between functions rather than at the end of the file.

Real malware authors often employ packing to obfuscate the entire binary. This work does not consider defense against packing obfuscation, except to note that the common defensive technique is to employ dynamic analysis. Our non-negative approach can be applied to the features derived from dynamic analysis as well, but beyond the scope of this paper. The possibility of evading non-negativity on dynamic features requires addressing the cat-and-mouse game around VM detection, stealthy malware, and the nature of features used. This discussion is important, but beyond the current ambit, which we limit to static analysis. We are interested here in whether or not non-negativity has benefit to the additive adversary, not more sophisticated ones.

\paragraph{Spam Threat Model Specifics}

For spam detection the adversary will be restrained to the insertion of new content into an existing spam message. This is because we are interested in the lower-effort "good word" attack scenario. Despite being less sophisticated, it remains effective today. Tackling wholly changed and newly crafted spam messages is beyond our current purview.

\paragraph{Image Threat model Specifics}

Image classification does not exhibit the asymmetric error costs that malware and spam do. The purpose of studying it is to determine if our non-negativity can have benefit to multi-class problems. It is intuitive that the answer would be "no," but we nevertheless find that some limited benefit exists. 

In this threat model, there is no "adding" or "removing" features, due to the intrinsic nature of images. As such we consider the $L_1$ distance between a original image $x$, and its adversarially perturbed counterpart $\hat{x}$. The adversary may arbitrarily alter any pixels, so long as $\|x-\hat{x}\|_1 < \epsilon$, where $\epsilon$ is a problem-dependent maximum distance.

\subsection{Attacking MalConv}\label{sec:malconv_exper}
 
MalConv is the primary focus of our interest, as gradient based attacks can not naively be applied to its architecture. Only recently have attacks been proposed \citep{Kolosnjaji2018,Kreuk2018}, and we will show that non-negativity allows us to thwart these adversaries. 
In MalConv, raw bytes of an executable are passed through a learned embedding layer which acts as a lookup table to transform each byte into an 8-dimensional vector of real values. This representation is then passed through a 1-dimensional gated convolution, global max pooling, and then a fully connected layer with sigmoid output. To handle varying file sizes, all sequences of bytes are padded to a fixed length of 2,100,000\ using a special "End of File" value (256) from outside of the normal range of bytes (0--255). 

The raw bytes are both discrete and non-ordinal, which prevents gradient based attacks from manipulating them directly. \citet{Kreuk2018} (and independently \citet{Kolosnjaji2018}) devised a clever way of modifying gradient based attacks to work on EXEs, even with a non-differentiable embedding layer, and we will briefly recap their approach.  This is done by performing the gradient search of an adversarial example in the 8-dimensional vector space produced by the embedding layer. A perturbed vector is then mapped to the byte which produces the nearest neighbor in the embedding space. Keeping with the notation of \citeauthor{Kreuk2018}, let $\bm{M} \in \mathbb{R}^{n\times{}d}$ be the lookup table from the embedding layer such that $\bm{M}: \bm{X} \to \bm{Z}$ where $\bm{X}$ is the set of $n$ possible bytes and $\bm{Z} \subseteq \mathbb{R}^d$ is the embedding space. 
Then for some sequence of bytes $\bm{x} = (x_0, x_1, \ldots, x_L)$, we generate a sequence of vectors $\bm{z} = (\bm{M}[x_0], \bm{M}[x_1], \ldots, \bm{M}[x_L])$ were $\bm{M}[x_i]$ indicates row $x_i$ of $\bm{M}$. Now we generate a new vector $\bm{\widetilde{z}} = \bm{z} + \bm{\delta}$ where $\bm{\delta}$ is a perturbation generated from an adversarial attack. We map each element $\widetilde{z_i} \in \bm{\widetilde{z}}$ back to byte space by finding the nearest neighbor of $\widetilde{z_i}$ among the rows of $\bm{M}$. By applying this technique to only specific safe regions of a binary, the execution of gradient based attacks against MalConv are possible without breaking the binary. To ensure that a "safe" area exists, they append an unused section to the binary. The larger this appended section is, the more space the adversary has to develop a strong enough signal of "benign-ness" to fool the algorithm. 

We replicate the attack done by \citet{Kreuk2018} which uses the \textit{fast gradient sign method} (FGSM) \citep{Goodfellow2015} to generate an adversarial example in the embedding space. We find our $\bm{\widetilde{z}}$ by solving:
	$\bm{\widetilde{z}} = \bm{z} + \epsilon \cdot \sign \left(\nabla_{\bm{z}}\widetilde{\ell} \left( \bm{z},y;\bm{\theta} \right) \right)$, 
where $\widetilde{\ell}(\cdot)$ is the loss function of our model parameterized by $\bm{\theta}$ and $\bm{z}$ is the embedded representation of some input with label $y$. The new $\bm{\widetilde{z}}$ is then mapped back into byte space using the method previously discussed. We performed the attack on 1000 randomly selected malicious files, varying the size of the appended section used to generate the adversarial examples. 

For MalConv, adding an unused section allows an attacker to add benign features which overwhelm the classification. Our hypothesis is that \malconvP should be immune to the attack since it only learns to look for maliciousness, defaulting to a decision of benign when no other evidence is present. We also note that this corresponds well with how anti-virus programs prefer to have lower false positive rates to avoid interfering with users' applications. 

\subsection{Attacking N-Gram}\label{sec:ngram_exper}

The N-Gram model was trained using lasso regularized logistic regression on the top million most frequent 6-byte n-grams found in our 2 million file training set. The 6-byte grams are used as boolean features, where a 1 represents the n-gram's presence in a file. Lasso performed feature selection by assigning a weight of 0 to most of the n-grams. The resulting model had non-zero weights assigned to approximately 67,000 of the features.

We devise a white-box attack similar to the attack \citet{Kreuk2018} used against MalConv in that we inject benign bytes into an unused section appended to malicious files. Specifically, we take the most benign 6-grams by sorting them based on their learned logistic regression coefficients.  We add benign 6-grams one at a time to the malicious file until a misclassification occurs. This ends up being the same kind of approach \citet{Lowd2005a} used to perform "Good Word" attacks on spam filters, except we assume the adversary has perfect knowledge of the model. The simplicity of the N-Gram model allows us to do this targeted attack, and specifically look at the evasion rate as a function of the number of inserted features. 

To prevent these attacks, we train \ngramP using non-negative weight constraints on the same data. This model is prevented from assigning negative weights to any of the features. We also remove the lasso regularization from \ngramP as the constraints are already performing feature selection by pushing the weights of benign features to zero. 

\subsection{Spam Filtering}

As mentioned in the previous section, \citet{Lowd2005a} created "Good Word" attacks to successfully evade spam filters without access to the model. These attacks append common words from normal emails into spam in order to overwhelm the spam filter into thinking the email is legitimate.
In their seminal work, they noted that it was unrealistic to assume that an adversary would have access to the spam filter, and would thus need to somehow guess at which words are good words, or to somehow query the spam filter to steal information about which words are good. Others have simply used the most frequent words from the ham messages as a proxy to good word selection that an adversary could replicate~\citep{Jorgensen:2008:MIL:1390681.1390719,Zhou2007}. We take the more pessimistic approach that the adversary has full access to our model, and can simply select the words that have the largest negative coefficients (i.e. the most good-looking words) for their attack.
This is the same assumption we make in attacking the n-gram model.

By showing that our non-negative learning approach eliminates the possibility of good word attacks in this pessimistic case, we intrinsically cover all weaker cases of an adversary's ability. We note as well that \citeauthor{Lowd2005a} speculated the only effective solution to stop the Good Word attack would be to to periodically re-train the model. By eliminating the possibility of performing Good Word attacks, we increase the cost to operate for the adversary, as they must now exert more effort into crafting significantly novel spam to avoid detection. By eliminating the lowest-effort approach the adversary can take, we remediate a sub-component of the spam problem, but not spam as a whole. 

We train two logistic regression models on the TREC 2006 and 2007 Spam Corpora.\footnote{See \url{https://trec.nist.gov/data/spam.html}} The 2006 dataset contains 37,822 emails with 24,912 being spam. The 2007 dataset contains 75,419 messages with 50,199 of them being spam. We performed very little text preprocessing and represented each email as a vector of boolean features corresponding to the top 10,000 most common words in the corpus. The first model is trained with lasso regularization in a traditional manner. The second model is trained with non-negative constraints on the coefficients in order to isolate only the features predictive of spam during inference.

\subsection{Targeted Attacks on Image Classification}

For our image classification experiments we follow the recommendations of  \citet{Carlini:2017:AEE:3128572.3140444} for evaluating an adversarial defense. In addition to the FGSM attack, we will also use a stronger iterated gradient attack. Specifically we use the Iterated Gradient Attack (IGA) introduced in  \citep{Kurakin2017}, using Keras for our models and Foolbox \citep{Rauber2017} for the attack implementations. We evaluated the confidences at which such attacks can succeed against the standard and our non-negative models on MNIST, CIFAR 10 and 100, and Tiny ImageNet. 

We note explicitly that the IGA attack is not the most poweruful adversary we could use. Other attacks like Projected Gradient Decent (PGD) and the C\&W attack\citep{Carlini2017} are more successfully, and defeat our multi-class generalization of non-negative learning. We study IGA to show that there is some benefit, but that overall the multi-class case is a weakness of our approach. We find the results interesting and informative because our prior belief would have been that non-negativity would produce no benefit to the defender at all, which is not the case. 

We are specifically interested in defending against an adversary creating a high confidence targeted attack (e.g., a label was previously classified as "cat", but now is classified as "potato" with a probability of 99\%). As such we will look at the evasion rate for an adversary altering an image to other classes over a range of target probabilities $p$. The goal is to see the non-negative trained network have a lower evasion rate, especially for $p \geq 90\%$. 

For MNIST and CIFAR 10, since there are only 10 classes, we calculate the evasion rate at a certain target probability $p$ as the average rate at which an adversary can successfully alter the networks prediction to every other class and reach a minimum probability $p$. For CIFAR 100 and Tiny ImageNet, the larger number of classes prohibits this exhaustive pairwise comparison. Instead we evaluate the evasion rate against a randomly selected alternative class.

On MNIST, CIFAR 10, and CIFAR 100, due to their small image sizes ($\leq 32\!\times\!32$), we found that adversarial attacks would often "succeed" by changing the image to an unrecognizable degree. For this reason we set a threshold of 60 on the $L_1$ distance between the original image and the adversarial modification. If the adversarial modified image exceeded this threshold, we counted the attack as a failure. This threshold was determined by examining several images; more information can be found in the appendix.
For Tiny ImageNet this issue was not observed, and Foolbox's default threshold was used. 

\section{Results} \label{sec:results}

Having reviewed the method by which we will fight targeted adversarial attacks, and how the malware attacks will be applied, we will now present the results of our non-negative networks. First we will review those related to malware and spam detection, showing that non-negative learning effectively neutralizes evasion by a malware author. Then we will show how non-negative learning can improve robustness on several image classification benchmarks. 

\subsection{Malware Detection}\label{sec:malconv_result}

\begin{figure}[!h]
\centering
\begin{adjustbox}{max size={1.0\columnwidth}{0.85\textheight}}
\begin{tikzpicture}
  \begin{groupplot}[
      group style={
        	group name=myplot, group size=1 by 3,
			vertical sep=2.5cm,%
        },
        enlarge x limits=true,
      ]
    \centering

\nextgroupplot[
     title=Evasion Rate as Size of Appended Section Increases,
     legend style={at={(0.97,0.5)},anchor=east},
    ymax=120,
    xlabel=Appended Section Size as Percent of File,
    ylabel=Evasion Rate (\%),
    ytick={0, 20, 40, 60, 80, 100},
    ]
    \addplot +[red,mark=o,dashdotted,mark options={solid}] table [x=Percent, y=MalConv, col sep=comma] {CSVs/malconv_evade.csv};
    \addplot +[blue,mark=x,loosely dashed,mark options={solid}] table [x=Percent, y=MalConv+, col sep=comma] {CSVs/malconv_evade.csv};
    \legend{MalConv,MalConv\textsuperscript{+}}
    
    \nextgroupplot[
     title=Evasion Rate as Top Benign N-Grams are Added,
     legend style={at={(0.97,0.5),anchor=east}},
    ymax=120,
    xlabel=Number of Benign N-Grams Added,
    ylabel=Evasion Rate (\%),
    ytick={0, 20, 40, 60, 80, 100},
    ]
    \addplot +[red,mark=o,dashdotted,mark options={solid}] table [x=Percent, y=N-Gram, col sep=comma] {CSVs/ngram_evade.csv};
    \addplot +[blue,mark=x,loosely dashed,mark options={solid}] table [x=Percent, y=N-Gram+, col sep=comma] {CSVs/ngram_evade.csv};
    \legend{N-Gram,N-Gram+}

\end{groupplot}
\end{tikzpicture}
\end{adjustbox}
\caption{
Evasion rate (y-axis) for MalConv and N-Gram based models. 
Top
figure shows MalConv evasion as the appended section size increases, and 
bottom
figure shows the N-Gram evasion as the number of benign n-grams are added. 
The number of files that evade increase as the size of the appended section increases. The evasion rates remained fixed for all section sizes greater than 25\% of the file size.
}
\label{fig:malconv_evade}
\end{figure}
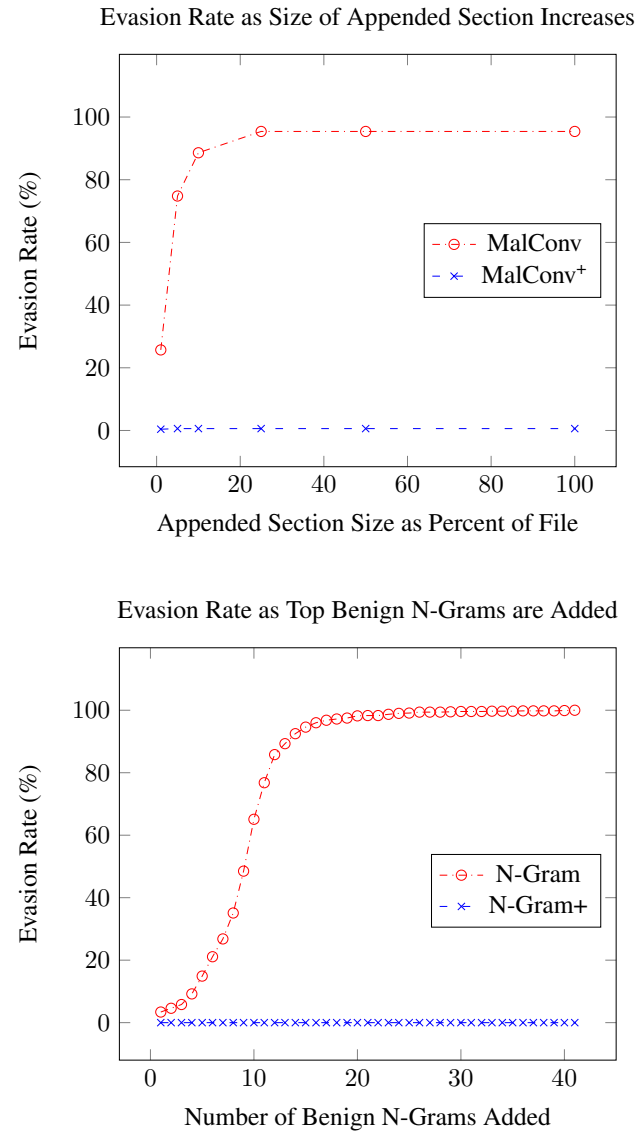

Using the method outlined in \autoref{sec:experiments}, \citet{Kreuk2018} reported a 100\% evasion rate of their model. As shown in \autoref{fig:malconv_evade}, our replication of the attack yielded similar results for MalConv, which
was evaded successfully for 95.4\% of the files. The other 4.6\% of files were all previously classified as malware with a sigmoid activation of 1.0 at machine precision
due to floating-point rounding. 
The attack fails for these cases since there is no valid gradient for this output. A persistent adversary could still create a successful adversarial example by replacing the sigmoid output with a linear activation function before running the attack.

Our non-negative learning provides an effective defense, with only 0.6\% of files able to evade \malconvP. Theoretically we would expect an evasion rate of 0.0\%. Investigating these successful evasions uncovered a hidden weakness in the MalConv architecture. We found that both MalConv and \malconvP learned to give a small amount of malicious predictive power to the special End of File (EOF) padding value. This is most likely a byproduct of the average malicious file size being less than the average benign file size in our training set, which causes the supposedly neutral EOF value itself to be seen as an indicator of maliciousness. 
The process of adding an unused file section necessarily reduces the amount of EOF padding tokens given to the network, as the file is increased in size (pushing it closer to the 2.1MB processing limit) the new section replaces the EOF tokens. Replacing the slightly malicious EOF tokens with benign content reduces the network's confidence in the file being malicious. 

The 0.6\% of files that evaded \malconvP only did so when files were small, and the appended section ended up comprising 50\% of the resulting binary. The slight maliciousness from the EOF was the needed feature to push the network into a decision of "malicious." However, the removal of EOFs by the unused section removed this slight signal, and pushed the decision back to "benign." 
If we instead replace the bytes of the unused section with random bytes from the uniform distribution, the files still evade detection. This means the evasion is not a function of the attack itself, but the modification of the binary that removes EOF tokens. 
A simple fix to this padding issue is to force the row of the embedding table corresponding to the special byte to be the zero vector during training. This would prevent the EOF token from providing any predictive power during inference.

We observed similar results for the N-Gram model. The evasion rate increases rapidly as benign features are added to the malicious files. We found that appending the top 41 most benign features resulted in a 100\% evasion rate. This attack is completely mitigated by \ngramP since none of its features have negative weights supporting the benign class. The only way to alter the classification would be to remove malicious n-grams from the files. Our results for both models are depicted in \autoref{fig:malconv_evade}.

\subsection*{Accuracy vs Defense}

The only drawback of this approach is the possible reduction in overall accuracy. Limiting the available information at inference time will likely reduce performance for most classification tasks. Alas, many security related applications exist because adversaries are present in the domain. We have shown that under attack our normal classifiers completely fail --- therefore a reduction in overall accuracy may be well worth the increase in model defensibility.
\autoref{tab:1} shows metrics from our models under normal conditions for comparison.

\begin{table}[tb]
\centering
\caption{Out of sample performance on malware detection in the absence of attack. }
\label{tab:1}
\begin{adjustbox}{max width=\columnwidth}
\begin{tabular}{@{}lcccc@{}}
\toprule
Classifier & Accuracy \% & Precision & Recall & AUC \%\\ \midrule
\textbf{MalConv}                    & 94.1    & 0.913    & 0.972    & 98.1 \\
\textbf{MalConv\textsuperscript{+}} & 89.4    & 0.908    & 0.888    & 95.3 \\
\textbf{N-Gram}                     & 95.5    & 0.926    & 0.987    & 99.6 \\
\textbf{N-Gram\textsuperscript{+}}  & 91.1    & 0.915    & 0.885    & 95.5 \\
\bottomrule
\end{tabular}
\end{adjustbox}
\end{table}

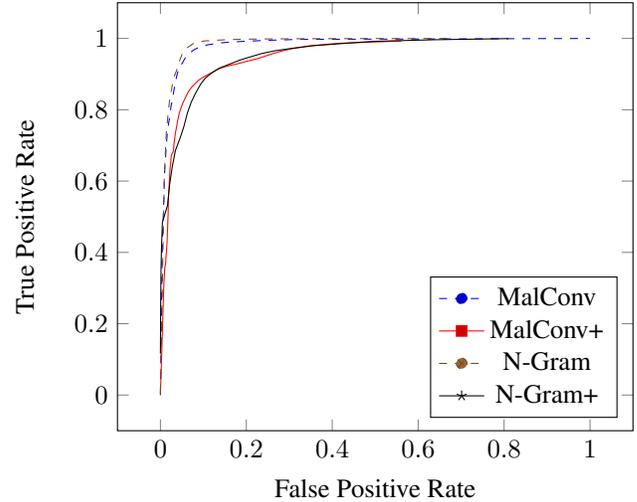
\begin{figure}[tb]
\centering
\vspace{1.0\baselineskip}
\begin{adjustbox}{max width=1.0\columnwidth}
\begin{tikzpicture}
\begin{axis}[
    xlabel={False Positive Rate},
    ylabel={True Positive Rate},
    legend pos=south east,
    ]
    \addplot+[dashed] table [mark=none,each nth point=73, x=fpr, y=tpr, col sep=comma] {CSVs/roc_malcon.csv};
    \addplot+[] table [mark=none,each nth point=163, x=fpr, y=tpr, col sep=comma] {CSVs/roc_malconP.csv};
    \addplot+[dashed] table [mark=none,each nth point=500, x=fpr, y=tpr, col sep=comma] {CSVs/roc_ngram.csv};
    \addplot+[] table [mark=none,each nth point=500, x=fpr, y=tpr, col sep=comma] {CSVs/roc_ngramP.csv};

    \legend{MalConv, MalConv+, N-Gram, N-Gram+}
    
\end{axis}
\end{tikzpicture}
\end{adjustbox}
\caption{
ROC curves for MalConv and N-Gram malware classifiers, with and without non-negative restraints, in the absence of attack.
}
\label{fig:malware_roc}
\end{figure}

Since different members of the security community have different desires with respect to the true positive vs. false positive trade-off, we also report the ROC curves for the MalConv, \malconvP, N-Gram, and \ngramP classifiers in \autoref{fig:malware_roc}.

While our non-negative approach has paid a penalty in accuracy, we note that we can see this has predominately come from a reduction in recall. Because features can only indicate maliciousness, some malicious binaries are labeled as benign due to a lack of information. This scenario corresponds to the preferred deployment scenario of security products in general, which is to have a lower false positive rate (benign files marked malicious) at the expense of false negatives (malicious files marked benign) 
\citep{Ferrand2016,Zhou:2008:MDU:1456377.1456393,learning-at-low-false-positive-rates}. 
As such the cost of non-negativity in this scenario is well aligned with its intended use case, making the cost palatable and the trade-off especially effective. 

To us it seems reasonable to accept a small loss in accuracy when not under attack in exchange for a large increase in accuracy when under attack. An interesting solution could be employing a pair of models, one constrained and one not, in addition to some heuristic indicating an adversarial attack is underway. The unconstrained model would generate labels during normal operations and fail over to the constrained model during attack. The confidence of the constrained model could be used as this switching heuristic as we empirically observe that the confidences during attack are much lower. 

Those who work in an industry environment and produce commercial grade AV products may object that our accuracy numbers do not reflect the same levels obtained today. We remind these readers that we do not have access to the same amount of data or access to the resources necessary to produce training corpora of similar quality, and so it should not be expected that we would obtain the same levels of accuracy as production systems. The purpose of this work is to show that a large class of models that have been used and attacked in prior works can be successfully defended against this common threat model. This comes at minor price, as just discussed, but this is the first technique to show that it can be wholly protective.

\subsection{Spam Filtering}

The accuracies for our traditional, unconstrained models were high on both datasets, but both were susceptible to our version of \citeauthor{Lowd2005a}'s "Good Word" attack. Both classifiers were evaded 100\% of the time by appending only 7 words to each message in the 2006 case and only 4 words in the 2007 case. These words correspond to the features with the lowest regression coefficients (i.e., negative values with high magnitude) for each model.

Use of the non-negative constraint lowers our accuracy for both datasets when not under attack, but \textit{completely eliminates} susceptibility to these attacks as all "Good Words" have coefficients of 0.0. The spam author would only be able to evade detection by removing words indicative of spam from their message. A comparison of performance is shown in \autoref{tab:2}.

\begin{table}[tbh]
\centering
\caption{Out of sample performance on spam filtering in the absence of attack.}
\label{tab:2}
\begin{adjustbox}{max width=\columnwidth}
\begin{tabular}{@{}lccccc@{}}
\toprule
Classifier & Accuracy \% & Precision & Recall  & AUC \% &F1 Score\\ \midrule
\textbf{2006 Lasso}        & 96.5    & 0.974   & 0.993  &97.1 & 0.983 \\
\textbf{2006 Non-Neg.} & 82.6    & 0.912   & 0.820  &83.5 & 0.864 \\
\textbf{2007 Lasso}        & 99.7    & 0.999   & 0.999  &99.7 & 0.999 \\
\textbf{2007 Non-Neg.} & 93.6    & 0.962   & 0.940  &93.0 & 0.951 \\
\bottomrule
\end{tabular}
\end{adjustbox}
\end{table}

\begin{figure*}[!h]
\centering
\begin{adjustbox}{max size={1.00\textwidth}{0.85\textheight}}
\begin{tikzpicture}
  \begin{groupplot}[
      group style={
        	group name=myplot,
            group size=4 by 1,
			vertical sep=2.0cm,%
        },
        enlarge x limits=true,
      ]
    \centering

\nextgroupplot[
	title=MNIST,
    legend pos=north east,
    ymax=30,
    xlabel=Target Confidence,
    ylabel=Evasion Rate,
	title style={font=\LARGE},
	label style={font=\Large},
    tick label style={font=\large},
    legend style={font=\large},
    ]
    \addplot +[red,mark=square,dashdotted,mark options={solid}] table [x=p, y=softmax_evade] {CSVs/mnist_fgsm.dat};
    \addplot +[red,mark=triangle,dashed,mark options={solid}] table [x=p, y=softmax_evade] {CSVs/mnist_iga.dat};
    \addplot +[blue,mark=o,dashdotted,mark options={solid}] table [x=p, y=nonneg_evade] {CSVs/mnist_fgsm.dat};
    \addplot +[blue,mark=otimes,dashed,mark options={solid}] table [x=p, y=nonneg_evade] {CSVs/mnist_iga.dat};
    \legend{Softmax FGSM, Softmax IGA, Non-Neg FGSM, Non-Neg IGA, style={font=\large}}

\nextgroupplot[
     title=CIFAR10,
     legend pos=north east,
    ymax=30,
    xlabel=Target Confidence,
	title style={font=\LARGE},
	label style={font=\Large},
    tick label style={font=\large},
    ]
    \addplot +[red,mark=square,dashdotted,mark options={solid}] table [x=p, y=softmax_evade] {CSVs/cifar10_target_fsgm.dat};
    \addplot +[red,mark=triangle,dashed,mark options={solid}] table [x=p, y=softmax_evade] {CSVs/cifar10_target_iga.dat};
    \addplot +[blue,mark=o,dashdotted,mark options={solid}] table [x=p, y=nonneg_evade] {CSVs/cifar10_target_fsgm.dat};
    \addplot +[blue,mark=otimes,dashed,mark options={solid}] table [x=p, y=nonneg_evade] {CSVs/cifar10_target_iga.dat};

\nextgroupplot[
     title=CIFAR100,
     legend pos=north east,
    ymax=8,
    xlabel=Target Confidence,
    xmode=log,
	title style={font=\LARGE},
	label style={font=\Large},
    tick label style={font=\large},
    ]
    \addplot +[red,mark=square,dashdotted,mark options={solid}] table [x=p, y=softmax_evade] {CSVs/cifar100_target_fgsm.dat};
    \addplot +[red,mark=triangle,dashed,mark options={solid}] table [x=p, y=softmax_evade] {CSVs/cifar100_target_iga.dat};
    \addplot +[blue,mark=o,dashdotted,mark options={solid}] table [x=p, y=nonneg_evade] {CSVs/cifar100_target_fgsm.dat};
    \addplot +[blue,mark=otimes,dashed,mark options={solid}] table [x=p, y=nonneg_evade] {CSVs/cifar100_target_iga.dat};

\nextgroupplot[
     title=Tiny ImageNet,
     legend pos=north east,
    ymax=8,
    xlabel=Target Confidence,
    xmode=log,
	title style={font=\LARGE},
	label style={font=\Large},
    tick label style={font=\large},
    ]
    \addplot +[red,mark=square,dashdotted,mark options={solid}] table [x=p, y=softmax_evade] {CSVs/mininet_fgsm.dat};
    \addplot +[red,mark=triangle,dashed,mark options={solid}] table [x=p, y=softmax_evade] {CSVs/mininet_iga.dat};
    \addplot +[blue,mark=o,dashdotted,mark options={solid}] table [x=p, y=nonneg_evade] {CSVs/mininet_fgsm.dat};
    \addplot +[blue,mark=otimes,dashed,mark options={solid}] table [x=p, y=nonneg_evade] {CSVs/mininet_iga.dat};

\end{groupplot}
\end{tikzpicture}
\end{adjustbox}
\caption{
Targeted evasion rate (y-axis) as a function of the desired misclassification confidence $p$ (x-axis) for four datasets. Due to the differing ranges of interest, right two figures shown in log scale for the x-axis. 
}
\label{fig:image_target_resist}
\end{figure*}
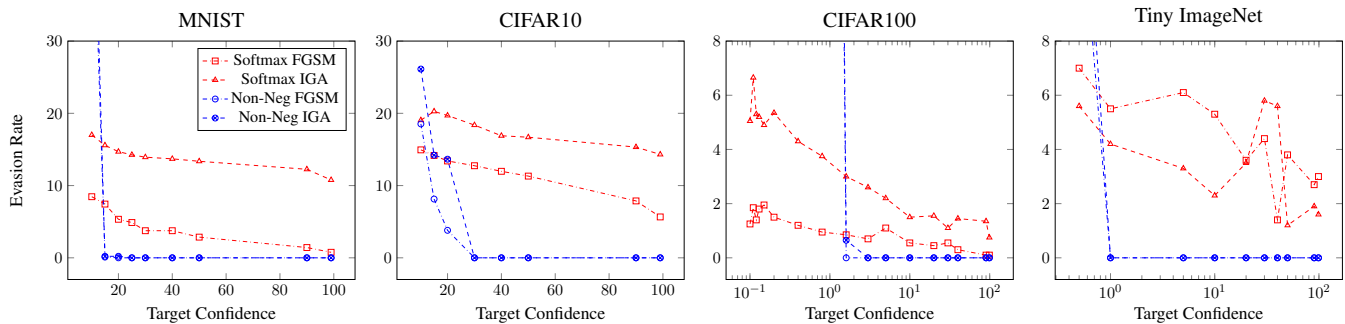

Despite the drops in accuracy imposed by our non-negative constraint, the results are better than prior works in defending against weaker versions of the "Good Word" attack. For example, \citet{Jorgensen:2008:MIL:1390681.1390719} developed a defense based on multiple instance learning. Their approach when attacked with all of their selected  good words had a precision of 0.772 and a recall of 0.743 on the 2006 TREC corpus. This was the best result of all their tested methods, but our non-negative approach achieves a superior 0.912 and 0.820 precision and recall respectively. While spam authors are not as restricted to modify their inputs, our approach forces them to move up to a more expensive threat model (removing and modifying features, rather than just adding) --- which we argue is of intrinsic value. 

\citet{Demontis2017} had concluded that there existed an "implicit trade-off between security and sparsity" in building a secure model in their Android malware work. At least for the additive adversary, we provide evidence with our byte n-grams and spam models that this is not an absolute. In both cases we begin with a full feature set and the non-negative approach learns a sparse model, where all "good words" (or bytes) are given coefficient values of zero. As such we see that sparsity and security occur together to defend against the additive adversary. 

\subsection{Image Classification}

Having investigated the performance of non-negative learning for malware detection, we now look at its potential for image classification. In particular, we find it is possible to leverage non-negative learning as discussed in \autoref{sec:image} to provide robustness against confident targeted attacks. That is to say if the predicted class is $y_i$, the adversary wants to trick the model into predicting class $y_j, j \neq i$ and that the confidence of the prediction be $\geq p$. 

For MNIST we will use LeNet. Our out of sample accuracy using a normal model is 99.2\%, while the model with non-negative constrained dense layers achieves 98.6\%.  For CIFAR 10 and 100 we use a ResNet based architecture.\footnote{v1 model taken from \url{https://tinyurl.com/keras-cifar10-restnet}} For CIFAR 10 we get 92.3\% accuracy normally, and 91.6\% with our non-negative approach. On CIFAR 100 the same architecture gets 72.2\% accuracy normally, and 71.7\% with our non-negative approach. For Tiny ImageNet, we also use a ResNet architecture with the weights of all but the final dense layers initialized pretrained from ImageNet.\footnote{ResNet50 built-in application from \url{https://keras.io/applications/\#resnet50}} The normal model has an accuracy of 56.6\%, and the constrained model 56.3\%.
The results as a function of the target confidence $p$ can be seen in \autoref{fig:image_target_resist}.

An interesting artifact of our approach is that the non-negative networks are easier to fool for low-confidence errors. We posit this is due to the probability distribution over classes becoming near uniform under attack. On CIFAR100 the y-axis is truncated for legibility since the evasion rate of FGSM is 93\% and IGA is 99\%. Similarly for non-negative Tiny ImageNet, FGSM and IGA achieve 14\% and 17\% evasion rates when $p=0.005$. 

Despite these initial high evasion rates, we can see in all cases the success of targeted adversarial attacks reaches 0\% as the desired probability $p$ increases. For MNIST and CIFAR10, which have only 10 classes, this occurs at up to a target 30\% confidence. As more classes are added, the difficulty of the attack increases. For Tiny ImageNet and CIFAR100, targeted attacks fail by $\leq 2\%$. 

If \textit{targeted} attacks from IGA were the only type of attack we needed to worry about, these results would also allow us to use the confidence as a method of detecting attacks. For example, CIFAR10 had the weakest results, needing a target confidence of 30\% before targeted attacks failed. The average predicted confidence of the non-negative network on the test set was 93.8\%. This means we can use the confidence itself as a measure of network robustness. If we default to a "no-answer" for everything with a confidence of 40\% or less on CIFAR10, and assume anything below that level is an attack and error, the accuracy would have only gone down 1.2\%.

In order to determine if non-negative constraints are merely acting as a gradient obfuscation technique \citep{Athalye2018}, we also attempted a black box attack by attacking a substitute model without the non-negative constraints and assessing whether the perturbed images created by this attack were able to fool the constrained model. In order to make the attack as strong as possible, the unconstrained network was the same as the network that was used to warm-start the non-negative training. This should maximize the transferability of attacks from one to the other. Despite this similarity, transfered attacks had only a 1.042\% success rate, which is one reason we believe that non-negative constraints are not merely a form of gradient obfuscation.

We emphasize that these results are evidence that we can extend non-negativity to provide benefit in the multi-class case. Our approach appears to have lower cost in multi-class case than in the binary, as accuracy drops by less than 1\% in each dataset. While the cost is lower, its utility is lower as well. Our multi-class non-negative approach provides no benefit in \textit{untargeted} attacks --- where any error by the model is acceptable to the attacker --- even under the weaker FGSM attack. When even stronger attacks like Projected Gradient Descent are used, our approach is also defeated in the targeted scenario. Under the moderate-strength IGA attack, we also see that susceptibility to evasion is increased for low-confidence evasions. In total, we view these results as indicative that non-negativity can have utility for the multi-class case and provide some level of benefit that is intrinsically interesting, but more work is needed to determine a better way to apply the technique. 

\section{Conclusion} \label{sec:conclusion}

We have shown that an increased robustness to adversarial examples can be achieved through non-negative weight constraints. Constrained binary classifiers can only identify features associated with the positive class during test time. Therefore, the only method for fooling the model is to remove features associated with that class. This method is particularly useful in security-centric domains like malware detection, which have well-known adversarial motivation. Forcing adversaries to remove maliciousness in these domains is the desired outcome. We have also described a technique to generalize this robustness to multi-class domains such as image classification. We showed a significant increase in robustness to targeted adversarial attacks while minimizing the amount of accuracy lost in doing so.

\bibliography{Mendeley,Will}

\begin{thebibliography}{}

\bibitem[\protect\citeauthoryear{Athalye, Carlini, and
  Wagner}{2018}]{Athalye2018}
Athalye, A.; Carlini, N.; and Wagner, D.
\newblock 2018.
\newblock {Obfuscated Gradients Give a False Sense of Security: Circumventing
  Defenses to Adversarial Examples}.
\newblock In {\em International Conference on Machine Learning (ICML)}.

\bibitem[\protect\citeauthoryear{Biggio, Fumera, and Roli}{2014}]{Biggio2014}
Biggio, B.; Fumera, G.; and Roli, F.
\newblock 2014.
\newblock {Security evaluation of pattern classifiers under attack}.
\newblock {\em IEEE Transactions on Knowledge and Data Engineering}
  26(4):984--996.

\bibitem[\protect\citeauthoryear{Carlini and
  Wagner}{2017a}]{Carlini:2017:AEE:3128572.3140444}
Carlini, N., and Wagner, D.
\newblock 2017a.
\newblock {Adversarial Examples Are Not Easily Detected: Bypassing Ten
  Detection Methods}.
\newblock In {\em Proceedings of the 10th ACM Workshop on Artificial
  Intelligence and Security}, AISec '17,  3--14.
\newblock New York, NY, USA: ACM.

\bibitem[\protect\citeauthoryear{Carlini and Wagner}{2017b}]{Carlini2017}
Carlini, N., and Wagner, D.
\newblock 2017b.
\newblock {Towards Evaluating the Robustness of Neural Networks}.
\newblock In {\em 2017 IEEE Symposium on Security and Privacy (SP)},  39--57.
\newblock IEEE.

\bibitem[\protect\citeauthoryear{Chistyakov \bgroup et al\mbox.\egroup
  }{2017}]{Chistyakov2017}
Chistyakov, A.; Lobacheva, E.; Kuznetsov, A.; and Romanenko, A.
\newblock 2017.
\newblock {Semantic Embeddings for Program behavior Patterns}.
\newblock In {\em ICLR Workshop}.

\bibitem[\protect\citeauthoryear{Chollet}{2015}]{chollet2015keras}
Chollet, F.
\newblock 2015.
\newblock {Keras}.

\bibitem[\protect\citeauthoryear{Chorowski and Zurada}{2015}]{Chorowski2015}
Chorowski, J., and Zurada, J.~M.
\newblock 2015.
\newblock {Learning Understandable Neural Networks With Nonnegative Weight
  Constraints}.
\newblock {\em IEEE Transactions on Neural Networks and Learning Systems}
  26(1):62--69.

\bibitem[\protect\citeauthoryear{Dalvi \bgroup et al\mbox.\egroup
  }{2004}]{Dalvi:2004:AC:1014052.1014066}
Dalvi, N.; Domingos, P.; {Mausam}; Sanghai, S.; and Verma, D.
\newblock 2004.
\newblock {Adversarial Classification}.
\newblock In {\em Proceedings of the Tenth ACM SIGKDD International Conference
  on Knowledge Discovery and Data Mining}, KDD '04,  99--108.
\newblock New York, NY, USA: ACM.

\bibitem[\protect\citeauthoryear{Demontis \bgroup et al\mbox.\egroup
  }{2017}]{Demontis2017}
Demontis, A.; Melis, M.; Biggio, B.; Maiorca, D.; Arp, D.; Rieck, K.; Corona,
  I.; Giacinto, G.; and Roli, F.
\newblock 2017.
\newblock {Yes, Machine Learning Can Be More Secure! A Case Study on Android
  Malware Detection}.
\newblock {\em IEEE Transactions on Dependable and Secure Computing}  1--1.

\bibitem[\protect\citeauthoryear{Ferrand and Filiol}{2016}]{Ferrand2016}
Ferrand, O., and Filiol, E.
\newblock 2016.
\newblock {Combinatorial detection of malware by IAT discrimination}.
\newblock {\em Journal of Computer Virology and Hacking Techniques}
  12(3):131--136.

\bibitem[\protect\citeauthoryear{Goodfellow, Shlens, and
  Szegedy}{2015}]{Goodfellow2015}
Goodfellow, I.~J.; Shlens, J.; and Szegedy, C.
\newblock 2015.
\newblock {Explaining and Harnessing Adversarial Examples}.
\newblock In {\em International Conference on Learning Representations (ICLR)}.

\bibitem[\protect\citeauthoryear{Grosse \bgroup et al\mbox.\egroup
  }{2016}]{DBLP:journals/corr/GrossePM0M16}
Grosse, K.; Papernot, N.; Manoharan, P.; Backes, M.; and McDaniel, P.~D.
\newblock 2016.
\newblock Adversarial perturbations against deep neural networks for malware
  classification.
\newblock {\em CoRR} abs/1606.04435.

\bibitem[\protect\citeauthoryear{Jorgensen, Zhou, and
  Inge}{2008}]{Jorgensen:2008:MIL:1390681.1390719}
Jorgensen, Z.; Zhou, Y.; and Inge, M.
\newblock 2008.
\newblock {A Multiple Instance Learning Strategy for Combating Good Word
  Attacks on Spam Filters}.
\newblock {\em J. Mach. Learn. Res.} 9:1115--1146.

\bibitem[\protect\citeauthoryear{Ko{\l}cz and Teo}{2009}]{citeulike:7099488}
Ko{\l}cz, A., and Teo, C.~H.
\newblock 2009.
\newblock {Feature Weighting for Improved Classifier Robustness}.
\newblock In {\em 6th Conference on Email and Anti-Spam (CEAS'09)}.

\bibitem[\protect\citeauthoryear{Kolosnjaji \bgroup et al\mbox.\egroup
  }{2018}]{Kolosnjaji2018}
Kolosnjaji, B.; Demontis, A.; Biggio, B.; Maiorca, D.; Giacinto, G.; Eckert,
  C.; and Roli, F.
\newblock 2018.
\newblock {Adversarial Malware Binaries: Evading Deep Learning for Malware
  Detection in Executables}.
\newblock In {\em 26th European Signal Processing Conference (EUSIPCO '18)}.

\bibitem[\protect\citeauthoryear{Kreuk \bgroup et al\mbox.\egroup
  }{2018}]{Kreuk2018}
Kreuk, F.; Barak, A.; Aviv-Reuven, S.; Baruch, M.; Pinkas, B.; and Keshet, J.
\newblock 2018.
\newblock {Adversarial Examples on Discrete Sequences for Beating Whole-Binary
  Malware Detection}.
\newblock {\em arXiv preprint}.

\bibitem[\protect\citeauthoryear{Kurakin, Goodfellow, and
  Bengio}{2017a}]{Kurakin2017AdversarialWorld}
Kurakin, A.; Goodfellow, I.; and Bengio, S.
\newblock 2017a.
\newblock {Adversarial examples in the physical world}.
\newblock In {\em International Conference on Learning Representations (ICLR)}.

\bibitem[\protect\citeauthoryear{Kurakin, Goodfellow, and
  Bengio}{2017b}]{Kurakin2017}
Kurakin, A.; Goodfellow, I.; and Bengio, S.
\newblock 2017b.
\newblock {Adversarial Machine Learning at Scale}.
\newblock In {\em International Conference on Learning Representations (ICLR)}.

\bibitem[\protect\citeauthoryear{Lowd and
  Meek}{2005a}]{Lowd:2005:AL:1081870.1081950}
Lowd, D., and Meek, C.
\newblock 2005a.
\newblock {Adversarial Learning}.
\newblock In {\em Proceedings of the Eleventh ACM SIGKDD International
  Conference on Knowledge Discovery in Data Mining}, KDD '05,  641--647.
\newblock New York, NY, USA: ACM.

\bibitem[\protect\citeauthoryear{Lowd and Meek}{2005b}]{Lowd2005a}
Lowd, D., and Meek, C.
\newblock 2005b.
\newblock {Good Word Attacks on Statistical Spam Filters}.
\newblock In {\em Conference on email and anti-spam (CEAS)},  125--132.

\bibitem[\protect\citeauthoryear{Maiorca, Biggio, and
  Giacinto}{2018}]{Maiorca2018}
Maiorca, D.; Biggio, B.; and Giacinto, G.
\newblock 2018.
\newblock {Towards Robust Detection of Adversarial Infection Vectors: Lessons
  Learned in PDF Malware}.
\newblock {\em arXiv preprint}.

\bibitem[\protect\citeauthoryear{Pascanu \bgroup et al\mbox.\egroup
  }{2015}]{export:249072}
Pascanu, R.; Stokes, J.~W.; Sanossian, H.; Marinescu, M.; and Thomas, A.
\newblock 2015.
\newblock {Malware Classification With Recurrent Networks}.
\newblock IEEE - Institute of Electrical and Electronics Engineers.

\bibitem[\protect\citeauthoryear{Raff and Nicholas}{2017}]{raff_shwel}
Raff, E., and Nicholas, C.
\newblock 2017.
\newblock {Malware Classification and Class Imbalance via Stochastic Hashed
  LZJD}.
\newblock In {\em Proceedings of the 10th ACM Workshop on Artificial
  Intelligence and Security}, AISec '17,  111--120.
\newblock New York, NY, USA: ACM.

\bibitem[\protect\citeauthoryear{Raff \bgroup et al\mbox.\egroup
  }{2016}]{raff_ngram_2016}
Raff, E.; Zak, R.; Cox, R.; Sylvester, J.; Yacci, P.; Ward, R.; Tracy, A.;
  McLean, M.; and Nicholas, C.
\newblock 2016.
\newblock {An investigation of byte n-gram features for malware
  classification}.
\newblock {\em Journal of Computer Virology and Hacking Techniques}.

\bibitem[\protect\citeauthoryear{Raff \bgroup et al\mbox.\egroup
  }{2018}]{MalConv}
Raff, E.; Barker, J.; Sylvester, J.; Brandon, R.; Catanzaro, B.; and Nicholas,
  C.
\newblock 2018.
\newblock {Malware Detection by Eating a Whole EXE}.
\newblock In {\em AAAI Workshop on Artificial Intelligence for Cyber Security}.

\bibitem[\protect\citeauthoryear{Rauber, Brendel, and
  Bethge}{2017}]{Rauber2017}
Rauber, J.; Brendel, W.; and Bethge, M.
\newblock 2017.
\newblock {Foolbox: A Python toolbox to benchmark the robustness of machine
  learning models}.

\bibitem[\protect\citeauthoryear{Russu \bgroup et al\mbox.\egroup
  }{2016}]{Russu:2016:SKM:2996758.2996771}
Russu, P.; Demontis, A.; Biggio, B.; Fumera, G.; and Roli, F.
\newblock 2016.
\newblock {Secure Kernel Machines Against Evasion Attacks}.
\newblock In {\em Proceedings of the 2016 ACM Workshop on Artificial
  Intelligence and Security}, AISec '16,  59--69.
\newblock New York, NY, USA: ACM.

\bibitem[\protect\citeauthoryear{Sahs and Khan}{2012}]{Sahs2012}
Sahs, J., and Khan, L.
\newblock 2012.
\newblock {A Machine Learning Approach to Android Malware Detection}.
\newblock In {\em 2012 European Intelligence and Security Informatics
  Conference},  141--147.
\newblock IEEE.

\bibitem[\protect\citeauthoryear{Saxe and Berlin}{2015}]{Saxe2015}
Saxe, J., and Berlin, K.
\newblock 2015.
\newblock {Deep Neural Network Based Malware Detection Using Two Dimensional
  Binary Program Features}.
\newblock {\em CoRR abs/1508.03096}.

\bibitem[\protect\citeauthoryear{Suciu, Coull, and Johns}{2018}]{Suciu2018}
Suciu, O.; Coull, S.~E.; and Johns, J.
\newblock 2018.
\newblock {Exploring Adversarial Examples in Malware Detection}.
\newblock In {\em AAAI 2018 Fall Symposium Series: Adversary-Aware Learning
  Techniques and Trends in Cybersecurity (ALEC)}.

\bibitem[\protect\citeauthoryear{Szegedy \bgroup et al\mbox.\egroup
  }{2014}]{Szegedy2014}
Szegedy, C.; Zaremba, W.; Sutskever, I.; Bruna, J.; Erhan, D.; Goodfellow, I.;
  and Fergus, R.
\newblock 2014.
\newblock {Intriguing properties of neural networks}.
\newblock In {\em ICLR}.

\bibitem[\protect\citeauthoryear{Ugarte-Pedrero \bgroup et al\mbox.\egroup
  }{2016}]{Ugarte-Pedrero:2016:RRP:2976956.2976970}
Ugarte-Pedrero, X.; Balzarotti, D.; Santos, I.; and Bringas, P.~G.
\newblock 2016.
\newblock {RAMBO: Run-Time Packer Analysis with Multiple Branch Observation}.
\newblock In {\em Proceedings of the 13th International Conference on Detection
  of Intrusions and Malware, and Vulnerability Assessment - Volume 9721}, DIMVA
  2016,  186--206.
\newblock New York, NY, USA: Springer-Verlag New York, Inc.

\bibitem[\protect\citeauthoryear{Yih, Goodman, and
  Hulten}{2006}]{learning-at-low-false-positive-rates}
Yih, S. W.-t.; Goodman, J.; and Hulten, G.
\newblock 2006.
\newblock {Learning at Low False Positive Rates}.
\newblock In {\em Proceedings of the 3rd Conference on Email and Anti-Spam}.
\newblock CEAS.

\bibitem[\protect\citeauthoryear{Yuan \bgroup et al\mbox.\egroup
  }{2017}]{Yuan2017}
Yuan, X.; He, P.; Zhu, Q.; Bhat, R.~R.; and Li, X.
\newblock 2017.
\newblock {Adversarial Examples: Attacks and Defenses for Deep Learning}.
\newblock {\em arXiv}.

\bibitem[\protect\citeauthoryear{Zhou and
  Inge}{2008}]{Zhou:2008:MDU:1456377.1456393}
Zhou, Y., and Inge, W.~M.
\newblock 2008.
\newblock {Malware Detection Using Adaptive Data Compression}.
\newblock In {\em Proceedings of the 1st ACM Workshop on Workshop on AISec},
  AISec '08,  53--60.
\newblock New York, NY, USA: ACM.

\bibitem[\protect\citeauthoryear{Zhou, Jorgensen, and Inge}{2007}]{Zhou2007}
Zhou, Y.; Jorgensen, Z.; and Inge, M.
\newblock 2007.
\newblock {Combating Good Word Attacks on Statistical Spam Filters with
  Multiple Instance Learning}.
\newblock In {\em 19th IEEE International Conference on Tools with Artificial
  Intelligence(ICTAI 2007)},  298--305.
\newblock IEEE.

\end{thebibliography}

\newpage
\newpage
\begin{appendices}

\section{Threshold for CIFAR Adversaries}

When creating an adversarial attack, it is necessary that some portion of the image must be changed as an intrinsic part of the attack. There is currently considerable debate about the nature of how large that change ought to be, how it should be measured, and how much we should care about the nature of changes to the original image. All of these could be topics of research in their own right, and we do not claim to solve them in this work. 
We use the $L_1$ distance as a measure simply because it has been used by prior works, even if it is not ideal. 

We also take the stance that it is important that no perceptible difference between the input and attacked result is important, while recognizing that is not an agreed upon procedure by everyone in the community. Intuitively we find the lack of perceptible difference important because it leaves no ambiguity about the ground truth label. If the input is noticeably perturbed by an attack, the true label of the new attacked image may come into question. Our CIFAR 10 \& 100 results against the non-negative networks often produced large differences, brining us to a need to impose a threshold at which we will consider an attack a "failure."

\begin{figure*}[!h]
\begin{subfigure}{.49\textwidth}
  \centering
  \includegraphics[width=.9\columnwidth]{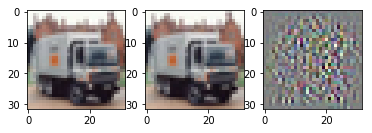}
  \caption{$L_1$ difference of 10, no perceptible difference.}
  \label{fig:sfig1}
\end{subfigure} 
\begin{subfigure}{.49\textwidth}
  \centering
  \includegraphics[width=.9\columnwidth]{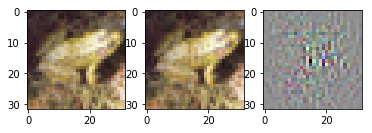}
  \caption{$L_1$ difference of 30, minute perceptible difference.}
  \label{fig:sfig2}
\end{subfigure}
\begin{subfigure}{.49\textwidth}
  \centering
  \includegraphics[width=.9\columnwidth]{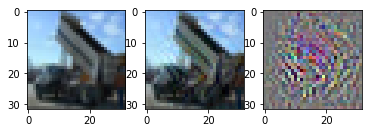}
  \caption{$L_1$ difference of 60, obvious perceptible difference, though objects appear mostly "the same". }
  \label{fig:sfig3}
\end{subfigure} 
\begin{subfigure}{.49\textwidth}
  \centering
  \includegraphics[width=.9\columnwidth]{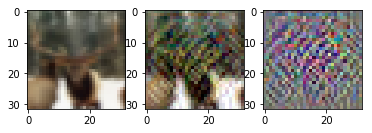}
  \caption{$L_1$ difference of 150, significant artifacts emerging in attack, original object is generally unrecognizable. }
  \label{fig:sfig4}
\end{subfigure}
\begin{subfigure}{.49\textwidth}
  \centering
  \includegraphics[width=.9\columnwidth]{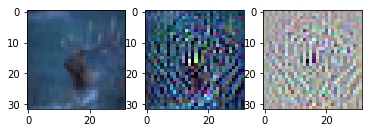}
  \caption{$L_1$ difference of 400, original object is no longer recognizable.}
  \label{fig:sfig5}
\end{subfigure} 
\begin{subfigure}{.49\textwidth}
  \centering
  \includegraphics[width=.9\columnwidth]{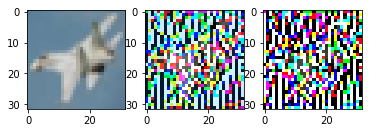}
  \caption{$L_1$ difference of 1000, the image has been completely destroyed.}
  \label{fig:sfig6}
\end{subfigure}
\caption{Examples of IGA attacks against CIFAR 10 images with our non-negative network. In each sub figure, the left most image is the original image, the middle is the attacked result, and the right shows the difference. Moving from sub-figure (a) to (f), the $L_1$ difference between the original and adversarial image increases. }
\label{fig:cifar_attacks_dists}
\end{figure*}

\autoref{fig:cifar_attacks_dists}
highlights the need to choose a threshold by providing examples of the spectrum of $L_1$ distances we observed between the original and the adversarial-generated images.
It starts with small distances, such as \autoref{fig:sfig1}, which has only an $L_1$ distance of 10 and is clearly still a truck. At the extreme end, we also had results like \autoref{fig:sfig6} which had an $L_1$ distance of 1000, and is wholly unrecognizable. We argue that such an attack must be a failure, as the input does not even resemble the distribution of images from CIFAR.

The question is, where does one draw the line? While we argue that an imperceptible difference is important to avoid label ambiguity, we have attempted to give deference in allowing large magnitude attacks while also recognizing that $L_1$ is not the ideal method to measure visual perceptual difference. As such we have experimentally selected a threshold of 60 as one that allows for perceptible differences, and at the edge of no longer being recognizable as its original class. 

Our decision to use a threshold of 60 is best shown in \autoref{fig:sfig3} where the $L_1$ difference starts to demonstrate an obvious perceptible difference. We feel this image represents a balance between the subjective ability of being able to still tell that it is a type of car / truck, and having difficulty recognizing what is in the image (or if it is still valid) without the context of the original image next to it. 

Allowing larger thresholds for the CIFAR attacks begins to enter a territory where it is not clear to us that the true label of the image has been retained. \autoref{fig:sfig4} shows one such example with a deer, where the adversarial image has the same colors but is unclear to us what the adversarial image should be labeled as.

\end{appendices}

\end{document}